# Deterministic Hypothesis Generation for Robust Fitting of Multiple Structures

Kwang Hee Lee, Chanki Yu, and Sang Wook Lee, *Member, IEEE*

**Abstract**— We present a novel algorithm for generating robust and consistent hypotheses for multiple-structure model fitting. Most of the existing methods utilize random sampling which produce varying results especially when outlier ratio is high. For a structure where a model is fitted, the inliers of other structures are regarded as outliers when multiple structures are present. Global optimization has recently been investigated to provide stable and unique solutions, but the computational cost of the algorithms is prohibitively high for most image data with reasonable sizes. The algorithm presented in this paper uses a maximum feasible subsystem (MaxFS) algorithm to generate consistent initial hypotheses only from partial datasets in spatially overlapping local image regions. Our assumption is that each genuine structure will exist as a dominant structure in at least one of the local regions. To refine initial hypotheses estimated from partial datasets and to remove residual tolerance dependency of the MaxFS algorithm, iterative re-weighted $L_1$ (IRL1) minimization is performed for all the image data. Initial weights of IRL1 framework are determined from the initial hypotheses generated in local regions. Our approach is significantly more efficient than those that use only global optimization for all the image data. Experimental results demonstrate that the presented method can generate more reliable and consistent hypotheses than random-sampling methods for estimating single and multiple structures from data with a large amount of outliers. We clearly expose the influence of algorithm parameter settings on the results in our experiments.

————————————  ————————————

## 1 INTRODUCTION

IN many computer vision problems observations or measurements are frequently contaminated with outliers and noise, and thus robust estimation is needed for model fitting. The "hypothesize-and-verify" framework is the core of many robust geometric fitting methods. The Random Sample Consensus (RANSAC) algorithm [24] is a widely used robust estimation technique, and most of the state-of-the-art methods are based on random sampling. They are comprised of two steps: (1) random hypotheses generation and (2) verification. These steps are performed iteratively. Many hypotheses of geometric model are randomly generated from minimal subsets of the data. The goal of random sampling is to generate many putative hypotheses for given geometric model. In the verification step, the hypotheses are evaluated according to robustness criterion to find the best model.

Random sampling-based methods have some drawbacks. In the majority of random sampling-based methods, the number of iterations needed to guarantee a desired confidence can be determined by a priori knowledge such as inlier ratio and inlier scale. For the single-structure data with unknown inlier ratio, it is crucial to determine the adequate number of iteration. The more heavily contaminated the data are, the lower the probability of hitting an all-inlier subset is. However, since the true inlier ratio is a priori unknown in many practical situations, it is necessary to be determined by users. The standard stopping criterion in RANSAC is based on an assumption that a model generated from an uncontaminated minimal subset is consistent with all-inliers. However, in practice, this assumption may be discrepant either increasing runtime or estimating incorrect solutions since inliers are perturbed by noise [18]. Furthermore, the existence of multiple structures makes the problem more difficult since the inliers belonging to other structures are regarded as outliers (pseudo-outliers). To the best of our knowledge, there is no stopping criterion to guarantee consistent and reliable results for multi-structure data.

When the number of iterations is insufficient, the random sampling-based techniques provide varying results for the same data and parameter settings. Despite their robustness, the random sampling-based methods provide no guarantee of consistency in their solutions due to the randomized nature [2]. Since many extensions of random sampling-based frameworks also follow the same heuristic of random sampling, none of them can guarantee deterministic solutions [2].

The various approaches to improve the efficiency of the random hypothesis generation for the estimation of single structure have been investigated [18, 29, 21, 22, 19, 20, 27]. They have been developed to increase the frequency of hitting all-inlier samples. Unfortunately, these methods are limited to single structure problem. In recent years, to deal with multiple structure data, some guided sampling methods [15, 16, 23] have been proposed. To perform guided sampling, a series of tentative hypotheses are generated from minimal subsets of the data in advance. Then, guided sampling based on preference analysis is performed. The quality of the initial hypotheses depends on the proportion of pseudo-outliers and gross outliers. Furthermore, since [15, 16, 23] have no clear stopping criterion, it is difficult to determine optimal number of iterations needed. Although the methods mentioned above improve the probability of hitting an all-inlier subsets in some ways, consistent performance cannot be guaranteed for the data with unknown inlier ratio or the number of structures. Many multiple-structure model fitting methods also start with random hypothesis generation [9, 10, 11, 13, 12, 14, 17, 8]. Some of these algorithms classify dataset based on randomly generated hypotheses and find model parameters [9, 10, 11]. Due to the same nature of random sampling, however, varying results may be produced from the same dataset.

Global optimization has recently been actively investigated for model fitting problems in computer vision [6, 2, 3, 25, 26]. Li has developed a global optimization method for the algebraic DLT (Direct Linear Transformation) problem that has fixed bounded variables [2]. He suggested an exact bilinear-MIP (Mixed Integer Programming) formulation and obtained globally optimal solutions using an LP (Linear Programming)-based BnB (Branch-and-Bound) algorithm. In [3], Yu et al. directly solved Big-M based MILP problem. While these methods guarantee globally optimal solution, high computational cost is required in general. For a large dataset, the global optimization methods require a great deal more running time than RANSAC. Furthermore, the presence of image features from multiple structures makes their computation cost even higher.

In this paper, we present a novel approach to reliable and consistent hypothesis generation for multiple-structure model fitting. Unlike previous random sampling methods, our method generates hypotheses using deterministic optimization techniques, and thus produces consistent results given a set of images. A whole image dataset is split into spatially overlapping circular regions of subsets, and a maximum feasible subsystem (MaxFS) problem is solved to generate consistent initial hypotheses in each region. Because of the reduction of data size using local regions, the MaxFS algorithm can generate the initial hypothesis from each local region with reasonable efficiency. Since the MaxFS algorithm yields a globally optimal solution only for the image subset in a spatially local region and the result is influenced by a residual tolerance value, an iterative reweighted $L_1$ (IRL1) minimization is carried out using all the data in the image to compensate for fitting errors from the subset and to get rid of the residual tolerance dependency. Our method is developed to find multiple structures under the assumption that a good hypothesis for each genuine structure will be found in at least one of the spatially local regions.

It may be noted that the use of spatial restriction for hypothesis generation is not unprecedented. There have been approaches to using spatial coherence in local regions for estimating single and multiple structures/motions [27, 28, 12]. They are based on random sampling and exploit spatial coherence mainly to increase the chance of finding all-inlier samples. Those methods therefore have the limitations of random sampling when the inlier ratio is unavailable. Our algorithm uses spatially overlapping local regions to generate the initial hypothesis efficiently and to exhaustively search for all genuine structures. It provides con-sistent hypotheses regardless of inlier ratio, noise effect and the number of structures.

Recently, in our previous work [36], a deterministic fitting method for multi-structure data has been proposed. To reduce high computation cost, the MaxFS algorithm is performed for small subset like the method presented in this paper. However, while initial subset is obtained by sorting keypoint matching scores in [36], the presented method does not require application-specific knowledge. In [36], there exist dependencies between hypotheses generated, since a sequential "fitting-and-removing" procedure is used. Therefore, it is impossible to generate hypotheses in parallel. On the other hand, the presented method can be immediately parallelized since there is no dependency between all of the hypotheses generated.

The rest of the paper is organized as follows: Section 2 introduces two deterministic methods for geometric fitting. Section 3 describes our algorithm based on MaxFS and IRL1 frameworks. Section 4 shows the experimental results on synthetic and real data, and we conclude in Section 5.

## 2 DETERMINISTIC METHODS FOR ROBUST GEOMETRIC FITTING

In this section, we briefly describe two main optimization techniques that we employ in our method.

### 2.1 Maximum Feasible Subsystem (MaxFS) Problem

The aim of a MaxFS algorithm is to find the largest cardinality set with constraints that are feasible [4, 2, 3]. In other words, it finds a feasible subsystem containing the largest number of inequalities for an infeasible linear system $\mathbf{Ax} \leq \mathbf{b}$ with real matrix $\mathbf{A} \in \Re^{k \times n}$, real vector $\mathbf{b} \in \Re^k$ and variable $\mathbf{x} \in \Re^n$. The objective of the MaxFS and RANSAC are the same. However, unlike the RANSAC, the MaxFS guarantees a global solution. The MaxFS problem admits the following mixed integer linear programming (MILP) formulation by introducing a binary variable $y_i$ for each of the inequalities:

$$\min_{\mathbf{x},\mathbf{y}} \sum_{i=1}^{k} y_i$$
$$\text{subject to } \sum_{j=1}^{n} a_{ij} x_j \leq b_i + M_i y_i, \quad \forall i \qquad (1)$$
$$\mathbf{x} \in \Re^n, \quad y_i \in \{0,1\}, \quad i=1,\dots,k$$

where $M_i$ is a large positive number that converts an infeasible inequality into a feasible one when $y_i = 1$. The case where $y_i = 0$ indicates that the inequality is feasible. Note that if $y_i = 1$, then the $i^{th}$ constraint is automatically deactivated. This MILP formulation is known as the Big-M method [1].

Generally, MILP problems are solved using the LP-based BnB (Linear Programming-based Branch and Bound) or LP-based BnC (LP-based Branch and Cut) methods. LP-based BnB and BnC guarantee global optimality of their solutions [3, 30]. The MILP problem is expensive in terms of computational cost, which is known to be NP-hard. Thus, only relatively small problems can be solved practically. While the exact Big-M MILP formulation is useful for small models, it is not effective on larger models for reasons of computational inefficiency [1]. To solve the geometric fitting problem including heavily contaminated multi-structure data, MaxFS demands a much higher computational cost than RANSAC style methods.

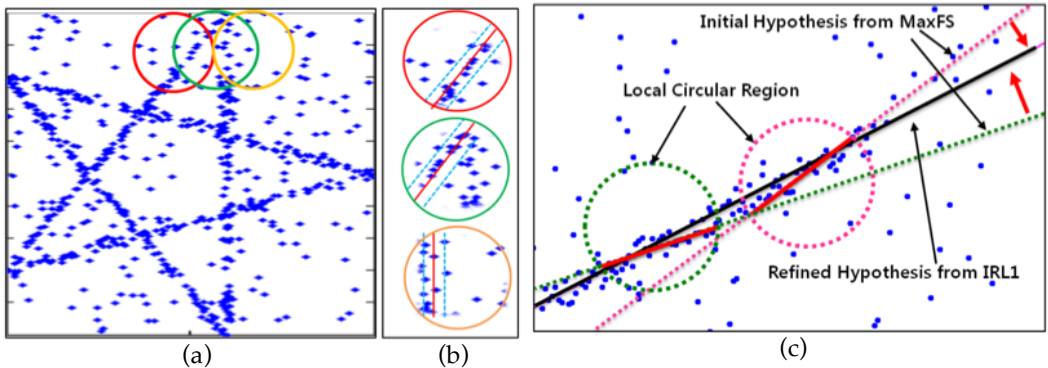

Fig. 1. Overview of our approach. (a) Overlapping circular regions for hypothesis generation. (b) Detection of dominant structures in local regions. (c) IRL1 minimization for refining initial hypothesis.

## 2.2 Iterative Re-weighted L1 (IRL1) Minimization

The most common deterministic method for geometric fitting is the least-squares estimation. The best fit minimizes the sum of the squared residuals ($L_2$-norm). This method is optimal for Gaussian noise, but the algorithm mostly fails in the presence of outliers. Using the sum of the absolute residuals ($L_1$-norm) relatively brings about better results than using $L_2$-norm in the presence of outliers, since $L_1$ minimization puts less weight on large residuals than $L_2$ minimization. Nevertheless, $L_1$ minimization still cannot guarantee robustness in the case of severely contaminated data or multiple structured data with outliers because the influence function has no cut off point. Although these methods always have a global solution, reliable output cannot be guaranteed in the presence of severe outliers.

Iterative re-weighted $L_1$ (IRL1) minimization has been presented by Candès, Wakin and Boyd [31]. The IRL1 algorithm solves a sequence of weighted $L_1$-norm problems where the weights are determined according to the estimated coefficient magnitude. The IRL1 minimization algorithm is summarized in Algorithm 1. $\mathbf{W}$ is a diagonal weighting matrix from the $t^{th}$ iteration with $i^{th}$ diagonal element $w_i^t$ and $a$ is a stability parameter which affects the stability of the algorithm.

In [31], experimental results show that it often outperforms standard $L_1$ minimization. Although each iteration of the algorithm solves a convex optimization problem, the overall algorithm does not. Therefore, one cannot expect this algorithm to always find a global minimum. Consequentially, it is important to determine a good starting point for the algorithm. In [31], initializing with the solution to standard $L_1$ minimization has been introduced. However, it is unsuitable for problems with heavily contaminated multi-structure data.

---

**Algorithm 1.** IRL1 algorithm [31]

**Input:** $\mathbf{y} = \mathbf{Ax}$

**Output:** $\mathbf{x}^t$

1:    Initialize: Set the weights $w_i^0 = 1$ for $i = 1\dots d$

2:    **Repeat**

3:        $t = t+1$

4:        $\mathbf{x}^t = \arg\min_{\mathbf{x}} |\mathbf{W}^t \mathbf{x}|_1, \; s.t. \; \mathbf{y} = \mathbf{Ax}$

5:        $w_i^{t+1} = \dfrac{1}{|x_i^t| + a}$

6:    **Until** convergence or a fixed number of times

## 3 HYPOTHESIS GENERATION USING DETERMINISTIC OPTIMIZATION TECHNIQUES

We present a deterministic algorithm to generate a reliable and consistent hypothesis for robust fitting of multiple structures. The whole data space is split into densely overlapping circular regions (see Fig. 1(a)). In each region, an initial hypothesis is generated from the maximum inliers. However, the initial hypothesis may be slightly biased even if it is generated from pure inliers, since the estimated inliers are from the local region. To estimate the best fit for each genuine structure from the initial hypothesis, IRL1 minimization is performed for all of the data in the image (see Fig. 1(c)). The algorithm of our method is summarized in Algorithm 2.

**Algorithm 2.** MaxFS-IRL1 algorithm

**Input:** input data **D**, the number of data (in each subset) $k$, the fractional constant $\alpha$, $M_i$ (Big-$M$), the residual tolerance $\varepsilon$, the variance of weight function $\sigma$

**Output:** hypothesis sets $\{\boldsymbol{\Theta}^*_j\}_{j=1,\ldots l}$

1:     Split **D** into $l$ subset $\mathbf{D}_j$

2:     **For** $j = 1$ to $l$

3:        Estimate initial parameter $\ddot{\boldsymbol{\Theta}}^{MaxFS}$ using MaxFS from Eq. (6) (from $\mathbf{D}_j$)

4:        Initialize weights from $\ddot{\boldsymbol{\Theta}}^{MaxFS}$ from Eq. (7)

5:        Refine parameter $\boldsymbol{\Theta}^*_j$ using IRL1 minimization (from **D**) :

6:        **Repeat**

7:           Solve the weighted $L_1$ minimization problem from Eq. (8)

8:           Update the weights from Eq. (9)

9:        **Until** convergence or a fixed number of times

10:     **End**

### 3.1 Determination of Spatially Overlapping Circular Regions

Our algorithm splits a whole image region into many overlapping small circular regions. The input data (**D**) is the union of $l$ subsets

$$\mathbf{D} = \{\mathbf{x}_i\}_{i=1}^N = \bigcup_{j=1}^l \mathbf{D}_j, \quad (2)$$

where $\mathbf{D}_j$ is the set of data in the circular region $j$. Neighboring subsets, $\mathbf{D}_m$ and $\mathbf{D}_n$, share common data in the overlapping region ($\mathbf{D}_m \cap \mathbf{D}_n \neq \emptyset$).

It is assumed that every structure in the image appears as a dominant structure in at least one of the local regions, and thus it suffices for our algorithm to find only one dominant structure in a circular region. The remaining structures are missed due to the dominant structure's initial detection, yet they can be found in the other local regions. Figure 1 (b) shows an example where an undetected structure in a window (Fig. 1 (b), middle) becomes a dominant structure in a neighboring window (Fig. 1 (b), bottom).

The circles' sizes and positions should be determined depending on the number of data points to be included in the circular regions. The number of data points $k$ that a circle covers should be larger than the minimum number required for fitting a desired model for hypothesis generation. If $k$ becomes larger, the result is more reliable but the computational cost is higher. The smaller the interval between the circles is, the slimmer the chance of missing small structures becomes. To maintain even performance over the regions, the circular windows include approximately the same number of data points. Thus, the computational cost for initial hypothesis generation is about the same.

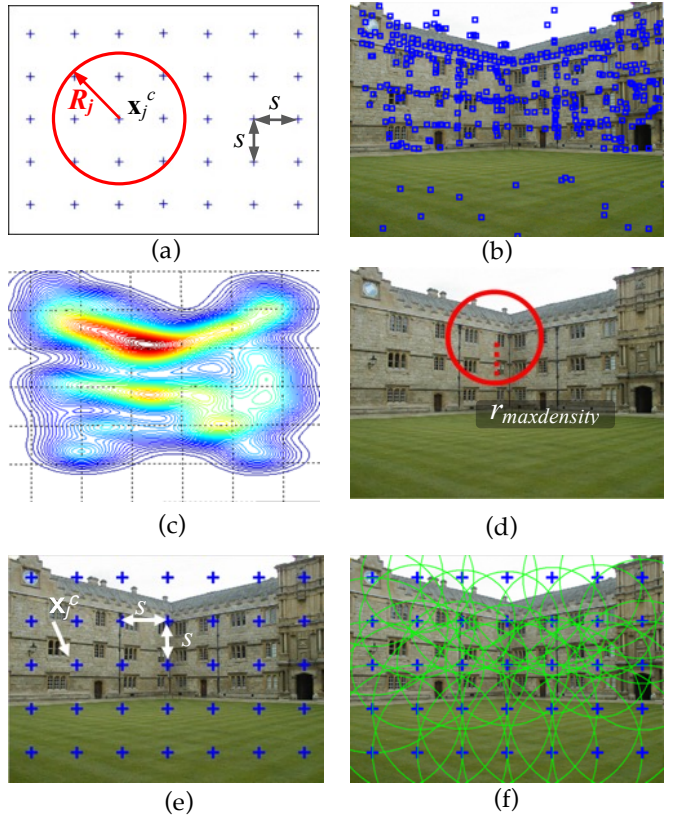

Fig. 2. (a) Center $\mathbf{x}_j^c$ and radius $R_j$ of circular region $j$ and step size $s$. (b) Input image and data points. (c) Data density map estimated from 2D KDE method. (d) Estimation of $r_{maxdensity}$. (e) Estimation of $s$ and $\mathbf{x}_j^c$. (f) Spatially overlapping circular regions.

Our algorithm places the centers of circular windows ($\mathbf{x}_j^c$s) at a regular interval $s$ in the horizontal and the vertical directions of image as illustrated in Fig. 2 (a), but the radii of the circles ($R_j$s) vary to keep the number of data points approximately the same. Once $k$ is set, the smallest circle that contains $k$ data points should appear where the data density is highest and we take its radius $r_{maxdensity}$ as a reference for determining the step size $s$ and $R_j$s for individual circles. We compute the density of data using the 2D Kernel Density Estimate (KDE) described in [7], and Figure 2 (c) shows the data density map for the image data points shown in Fig. 2(b). In Fig. 2 (d), the red circle represents the highest-density region with $k$-nearest neighbors and the dotted red line does $r_{maxdensity}$.

The step size $s$ is determined as follows:

$$s = r_{maxdensity}\alpha, \quad (3)$$

where $\alpha$ is a fractional constant. The subset $\mathbf{D}_j$ around the circle center $\mathbf{x}_j^c$ is defined as:

$$\mathbf{D}_j = \{\mathbf{x}_i \in \mathbf{D} \mid |\mathbf{x}_i - \mathbf{x}_j^c| \le R_j\},$$
$$R_j = \min(r_j, 2r_{maxdensity}), \quad (4)$$

where $r_j$ is the minimum raidus for the circle that contains $k$-nearest neighbors. Figure 2 (e) shows the center points of the circular regions (blue crosses), and an example of spatially overlapping circular regions (green circles) is shown in Fig. 2 (f).

Since $r_{maxdensity}$ is the shortest possible radius, $r_j$ is always longer than $r_{maxdensity}$ and thus substantial overlap between the adjacent circles is guaranteed. The constant $\alpha$ controls the extra degree of overlap. The maximum $R_j$ is set to $2r_{maxdensity}$ since $r_j$ becomes unmeaningfully long where data density is very low. The number of datapoints $k$ and the constant $\alpha$ are the most important parameters in our algorithm and we experimentally investigate the influence of their settings on the results in Secion 4.

### 3.2 Initial Hypothesis Generation using the MaxFS Algorithm

We determine the maximum feasible inliers in each local region. An initial hypothesis is generated from the maximum feasible inliers in each local region. A simple way of deterministically solving this problem would be to perform an exhaustive search. This is intractable due to the combinatorial explosion of subsets. On the other hand, random-sampling methods cannot guarantee consistent results due to their randomized nature and stopping criterion depends on the prior knowledge of information such as inlier ratio.

Unlike random-sampling methods, the MaxFS algorithm can guarantee maximum feasible inliers, even though the inlier ratio is unknown. Moreover, by splitting the problem into many small parts, the MaxFS algorithm can be performed quickly and efficiently. We use the algebraic Direct Linear Transformation (DLT) to estimate hypothesis parameters [5]. We can then formulate the DLT-based geometric fitting problem as a MaxFS problem. Each subset $\mathbf{D}_j$ is partitioned into the inlier-set $\mathbf{D}_j^I$ and the outlier-set $\mathbf{D}_j^O$ with $\mathbf{D}_j^I \subseteq \mathbf{D}_j$, $\mathbf{D}_j^O \subseteq \mathbf{D}_j$, $\mathbf{D}_j^I \cup \mathbf{D}_j^O = \mathbf{D}_j$ and $\mathbf{D}_j^I \cap \mathbf{D}_j^O = \emptyset$.

A maximum residual tolerance, $\varepsilon > 0$, provides a bound for the algebraic residual $d_i = |\mathbf{a}_i^T \Theta|$ at point $i$, where $\mathbf{a}_i^T$ is each row vector of $\mathbf{A}$ in the homogeneous equation $\mathbf{A}\Theta = \mathbf{0}$:

$$d_i = |\mathbf{a}_i^T \Theta| \le \varepsilon, \ \varepsilon > 0. \quad (5)$$

The MaxFS formulation of Eq. (5) is as follows:

$$\{\hat{\Theta}^{MaxFS}, \hat{\mathbf{y}}\} = \underset{\Theta, \mathbf{y}}{\operatorname{argmin}} \sum_{i=1}^{k} y_i$$
$$\text{subject to} \quad |\mathbf{a}_i^T \Theta| \le \varepsilon + M_i y_i, \ \forall i \quad (6)$$
$$\mathbf{c}^T \Theta = 1,$$
$$\Theta \in \mathfrak{R}^n, \ y_i \in \{0,1\}, i = 1,...,k.$$

where $M_i$ is a large positive number (called Big-M value).

The case where $y_i = 0$ indicates that the $i^{th}$ data is an inlier. If $y_i = 1$, the $i^{th}$ data is an outlier and the corresponding constraint is deactivated automatically. We use a linear constraint $\mathbf{c}^T \Theta = 1$, rather than the commonly used $\|\Theta\| = 1$, where $\mathbf{c}$ is a problem dependent vector determined by the user [5].

Note that our MaxFS algorithm solves Eq. (6) for every subset $\mathbf{D}_j$. Then, a series of hypotheses { $\hat{\Theta}_1^{MaxFS}$,…, $\hat{\Theta}_l^{MaxFS}$ } are generated from the maximum inlier-set where $l$ is the number of hypotheses.

In the hypothesis generation stage, although the MaxFS algorithm obtains a global solution for the local subset, the parameter vectors estimated in spatially restricted regions can be biased agaist the true structures. To refine initial hypotheses estimated from local datasets, IRL1 minimization is performed for all data in the image.

### 3.3 Hypothesis Refinement Using the IRL1 Minimization

In each local region, given an initial hypothesis generated from the MaxFS algorithm, initial weights of all the data can be determined as

$$w_i^{(0)} = \exp(\frac{-|r_i^{\hat{\Theta}^{MaxFS}}|}{\sigma}), i = 1,...,N, \quad (7)$$

where $r_i^{\hat{\Theta}^{MaxFS}}$ is the residual of data $x_i$ to the initial hypothesis and $\sigma$ controls the width of the weight function. Unlike the original IRL1 [31] minimization algorithm, our IRL1 minimization algorithm uses the MaxFS algorithm to generate the initial weights. Since the initial hypothesis generated from the MaxFS algorithm is robust to outliers, it may provide much better initial weights than the original IRL1 [31] minimization that uses the standard $L_1$ minimization as the method to generate initial weights.

After initial weights are set, the IRL1 minimization iteratively performs the two-step algorithm shown below. The first step solves the following weighted version of $L_1$ minimization:

$$\{\hat{\Theta}_t, \hat{\mathbf{y}}\} = \underset{\Theta, \mathbf{y}}{\operatorname{argmin}} \sum_{i=1}^{N} w_i^t y_i$$
$$\text{subject to} \quad |\mathbf{a}_i^T \Theta| \le y_i, \ \forall i \quad (8)$$
$$\mathbf{c}^T \Theta = 1,$$
$$\Theta \in \mathfrak{R}^n, \ y_i = [0,\infty], i = 1,...,N,$$

where $t$ is the index of current iteration.

In the next step, the weights of all the data are updated as follows:

$$w_i^{(t+1)} = \exp(\frac{-|r_i^{\hat{\Theta}_t}|}{\sigma}), \ i = 1,...,N, \quad (9)$$

where $r_i^{\hat{\Theta}_t}$ is the residual of data $x_i$ to the current hypothesis and $\sigma$ controls the width of the weight function. In each iteration, Eq. (8) and Eq. (9) are alternately performed (see

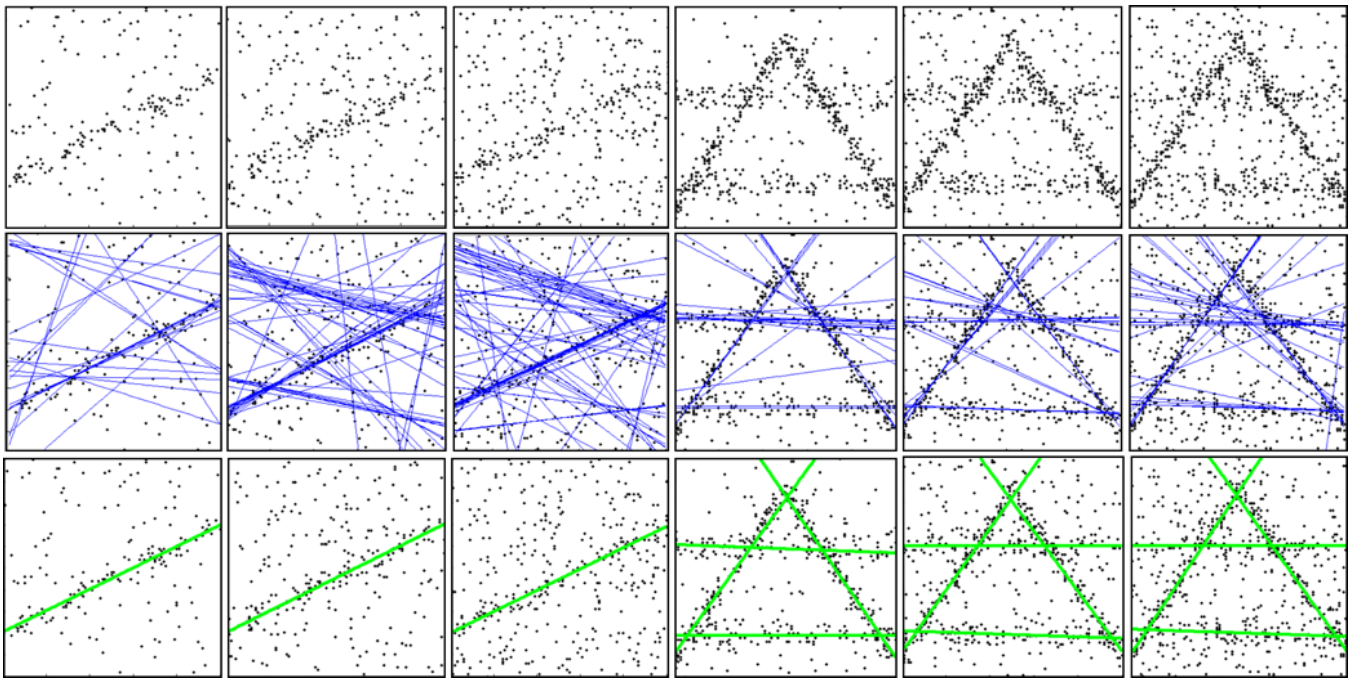

Fig. 3. Examples for line fitting. (Top row) Input data with different outlier ratio (50%, 66%, 75%, 20%, 33.3%, 42.8%). (Middle row) The generated hypotheses using the proposed method. (Bottom row) Fitting results.

Algorithm 2, 6~9). In other words, the current parameter vector $\hat{\Theta}_t$ is estimated by solving Eq. (8) and then weights of all the data are updated based on the current parameter vector by Eq. (9). The algorithm is repeated until convergence or for a fixed number of times.

Note that an advantage of IRL1 minimization is that the results are not influenced by residual tolerance, unlike with RANSAC or MaxFS. Therefore, our IRL1 minimization not only refines a hypothesis biased by local fitting, but also eliminates the residual tolerance dependency.

### 3.4 Complementary Role of MaxFS and IRL1

The hypotheses estimated from only MaxFS may deviate from the true structures since they are estimated in local regions. We use the IRL1 procedure to globally refine these hypotheses. The initial weights of the conventional IRL1 procedure are usually generated from the standard $L_1$ minimization which often fails when there are a large amount of outliers. We employ the MaxFS algorithm to generate much more reliable initial weights than the standard $L_1$ minimization. It is our intention to carefully combine the two algorithms to complement each other's limitations. The contribution of each component can be seen in Fig. 11 and Fig. 12.

## 4 EXPERIMENTAL RESULTS

We have implemented our algorithm in MATLAB using the LP/MILP solver GUROBI [30] which provides functions for the LP/MILP and a desktop with Intel i5-2500 3.30GHz (4 cores) and 3GB RAM was used for experiments. We used 4 cores only for solving mixed integer linear programming in each MaxFS problem. We measured the actual elapsed computation time and tested our proposed method on several synthetic and real datasets.

### 4.1 Experiments with Synthetic Datasets: Single and Multiple Line Fitting

The first set of results we show is produced from single and multiple 2D line fitting. We performed the DLT-based MaxFS and IRL1 algorithms for each data subset and the results are shown in Figs. 3 and 4. The residual tolerance $\varepsilon$ was set to 1 and the Big-M value of Eq. (6) was fixed to 10000. The variance of weight function $\sigma$ was fixed to 10. The number of data points in the circular region $k$ was fixed to 40. In the single line test, each line includes 100 inliers with Gaussian noise and various gross outliers. Noise level was fixed at 0.03, and the number of gross outliers varies from 100 to 300. In multiple line fitting tests, the same ratio of gross outliers and Gaussian noise was added. Final fitting from generated candidate hypotheses was performed using Li's method [8].

Figure 3 shows input synthetic data (Top row), corresponding hypotheses results (Middle row) and the final fitting results (Bottom row). The results show our algorithm generates good hypotheses from true structures. Even when the ratio of gross outliers increases, the proportion of good hypotheses is fairly consistent.

Figure 4 shows (First column) initial hypotheses from the MaxFS algorithm, (Second column) hypotheses after the first reweighted iteration, (Third column) hypotheses after the third reweighted iteration and (Last column) fitting results. We used the IRL1 minimization from all of the data to reduce the errors from subset data. In the first row, noise level was fixed at 0.01 and the number of gross outliers was fixed at 200. In the second row, noise level is fixed at 0.03 and the number of gross outliers was the same. With the residuals of the whole dataset being used for estimation, hypotheses are refined for genuine structures.

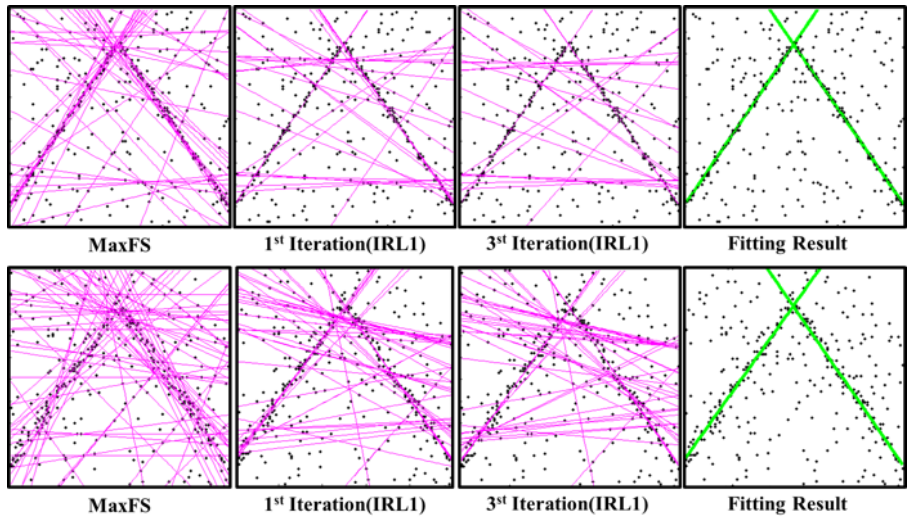

Fig. 4. Hypothesis refinement through reweighted L1 iterations. (First column) Initial hypotheses from Max-FS framework. (Second column) Hypotheses after the first reweighted iteration. (Third column) Hypotheses after the third reweighted iteration. (Last column) Fitting results.

## 4.2 Experiments with Real Datasets

We tested four methods including our MaxFS-IRL1 algorithm on several real datasets. For performance evaluation and comparison, we measured the actual elapsed computation time. Images and keypoint correspondences were acquired from the Oxford VGG dataset [32] and the Adelaide RMF dataset [34, 35]. The image pairs used in the experiment are shown in Figs. 5 and 6. Yellow crosses indicate the gross outliers randomly generated and other colored squares indicate the inliers of each structure. Each dataset included the various rates of outliers. We used manually labeled keypoint correspondences which were obtained by SIFT matching. Given true inlier-sets and the number of structures, for each structure, a single hypothesis having minimum re-projection errors between ground truth inliers and estimated hypothesis parameters was selected. The quality of hypotheses estimated was evaluated by averaging re-projection errors for all structures.

### 4.2.1 Analysis: MaxFS-IRL1 Framework

We performed the MaxFS-IRL1 method to estimate planar homography and affine fundamental matrix. For our MaxFS-IRL1 algorithm, the Big-M value in Eq. (6) was set to 10000 for both applications. For estimation we used Direct Linear Transformation (DLT) and the residual was taken as the Sampson distance [5].

We experimentally examined the effects of parameter $k$ (subset size) and $\alpha$ (fractional constant) on different datasets with 70% of outliers. Firstly, we investigated the effects of $k$ in the range of [10, 50] with fixed $s$=100 for the homography estimation and $s$=50 for the affine fundamental matrix estimation. Figs. 7(a) and 7(b) show the re-projection errors and computation time for the homography estimation with the MaxFS-IRL1 method, respectively. Only the results from three datasets are shown in the plots. Figs. 8(a) and 8(b) show the re-projection errors and the computation time for the affine fundamental matrix estimation, respectively. It can be seen in Figs. 7 and 8 that high accuracy is achieved for the $k$ from about 30 and above for the homography estimation and 20 and above for the affine fundamental matrix estimation, and the computation time gradually increases with $k$. For the MaxFS-IRL1 algorithm, we set $k$ to 30 for the homography estimation and 20 for the affine fundamental matrix estimation to attain both accuracy and computational efficiency. Secondly, we investigated the effects of $\alpha$ in the range of [0.25, 1.5] with predetermined $k$. Figs. 7(c) and 7(d) show the re-projection errors and computation time for the homography estimation with the MaxFS-IRL1 method, respectively. Figs. 8(c) and 8(d) show the re-projection errors and the computation time for affine fundamental matrix estimation with the MaxFS-IRL1 method, respectively. Based on these results, we set $\alpha$ to 1 for the homography estimation and 0.5 for the affine fundamental matrix estimation by considering both accuracy and computational efficiency.

We empirically examined the effects of parameters $\varepsilon$ (residual tolerance) and $\sigma$ (variance of weight function). Firstly, we investigated the effect of $\varepsilon$ with fixed $\sigma$=1 in the range of [0.1, 0.3, … 2.9] for the image pairs for homography estimation (Fig. 5) and in the range of [0.0005, 0.001, …, 0.005] for the image pairs for affine fundamental matrix estimation (Fig. 6). Figs. 9(a) and 9(b) show the re-projection errors from the MaxFS algorithm and MaxFS-IRL1 method for one of several structures in the CollegeIII data with 60% of the outliers included and the Carchipscube data with 40% of the outliers included. The results clearly show that our algorithm is stable within a wide range of the inlier tolerance $\varepsilon$. We set $\varepsilon$=0.5 for homography estimation and $\varepsilon$=0.001 for affine fundamental matrix estimation based on our test results. Secondly, while $\sigma$ is varied in the range of [1, 2, … 5] with fixed $\varepsilon$, we recorded the re-projection error in each iteration. Figs. 9(c) and 9(d) shows the results for the Neem data with 50% of the outliers included and the Cubebreadtoychips data with 30% of the outlier included. In most experiments, when $\sigma$<5, the results were stable and accurate. Considering the convergence speed and accuracy, we set $\sigma$=3 by default for both applications.

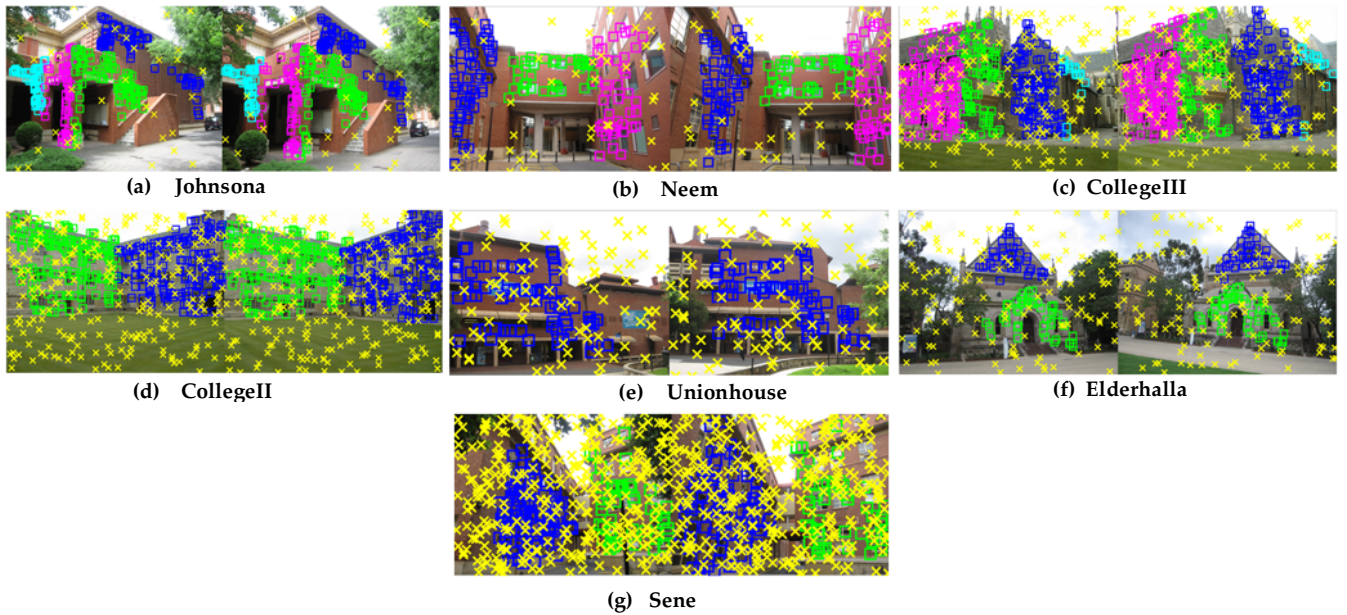

Fig. 5. Test image-sets used for homography estimation experiment. Yellow crosses indicate the gross outliers randomly generated and other colored squares indicate the inliers of each structure.

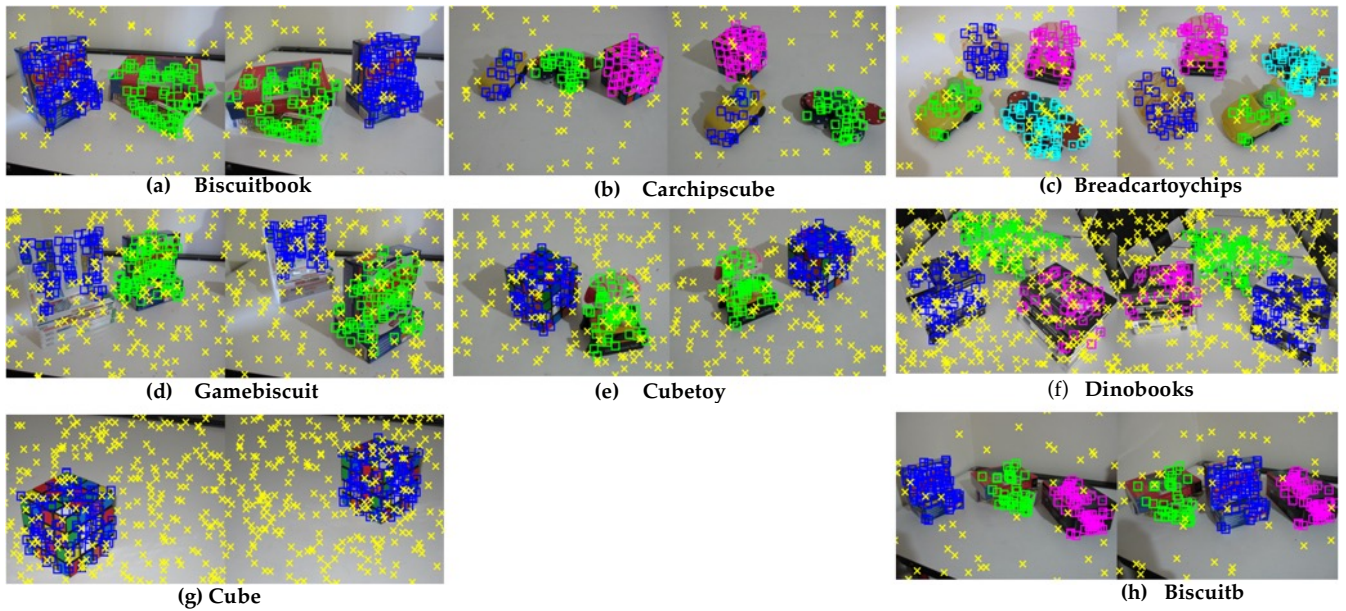

Fig. 6. Test image-sets used for affine fundamental matrix estimation experiment. Yellow crosses indicate the gross outliers randomly generated and other colored squares indicate the inliers of each structure.

### 4.2.2 Comparison with Other Methods

Our algorithm was compared with uniform random sampling (Random Sampling) [24, 33], NAPSAC [27] and the state-of-the-art algorithm Multi-GS [23, 34]. We implemented NAPSAC in MATLAB. For performance evaluation, we measured elapsed computation time and the number of generated hypotheses (L) and computed the re-projection errors (maximum error and standard deviation). The results for the four algorithms are summarized in Tables 1 and 2 with the best results shown in **bold**.

For Random Sampling, NAPSAC and Multi-GS, 50 random runs were performed. Unlike these competitors, since our method runs untill completion of the algorithm, the elapsed computation time for our method was not limited but measured. To ensure a fair comparison, three competing methods were performed for similar periods of time (elapsed computation time) along with our algorithm. In this experiment, we selected all of the parameters based on section 4.2.1 for our MaxFS-IRL1 framework.

**Homography Estimation:** We tested the performance of our proposed method for estimating planar homographies on real image data. For the homography estimation, the Big-M value of Eq. (6) was fixed to 10000, the residual tolerance $\varepsilon$ was fixed to 0.5, the variance of weight function $\sigma$ was fixed to 3, the number of iterations in IRL1 was fixed to 5, fractional constant α was fixed to 1 and the number of data points in the circular region $k$ was fixed to 30.

Table 1 summarizes the performance of four methods for estimating planar homography for seven datasets (see Fig. 5. (a-g)) with various outlier ratios. The results demonstrate that our method yields more reliable and consistent results with reasonable computational efficiency.

**Affine Fundamental Matrix Estimation:** We also tested the performance of our proposed method for estimating an affine fundamental matrix on real image data. For the affine fundamental matrix estimation, the Big-M value was fixed to 10000, the residual tolerance ε was fixed to 0.001, the variance of weight function σ was fixed to 3, the number of iterations in IRL1 also was fixed to 5, fractional constant α was fixed to 0.5 and the number of data points in the circular region $k$ was fixed to 20.

hypothesis is reliable but may not yield the minimum errors for true inliers. Note that random sampling-based methods produced large variations in their results.

### 4.2.3 Performance under Increasing Outlier Rates

We compared the performance of four algorithms under different outlier rates. We increased the outlier ratio from 10% to 80% for the CollegeIII data and the Biscuitbookbox data. Figs. 10(a) and 10(c) show the re-projection errors produced by the four methods on the two test datasets as the outlier ratio increases and Figs. 10(b) and 10(d) show the corresponding standard deviations for the re-projection errors. Our algorithm outperforms the other algorithms when outlier ratio is high over similar periods of time (elapsed computation time). Since the probability of producing an all-inlier subset decreases with random sam-

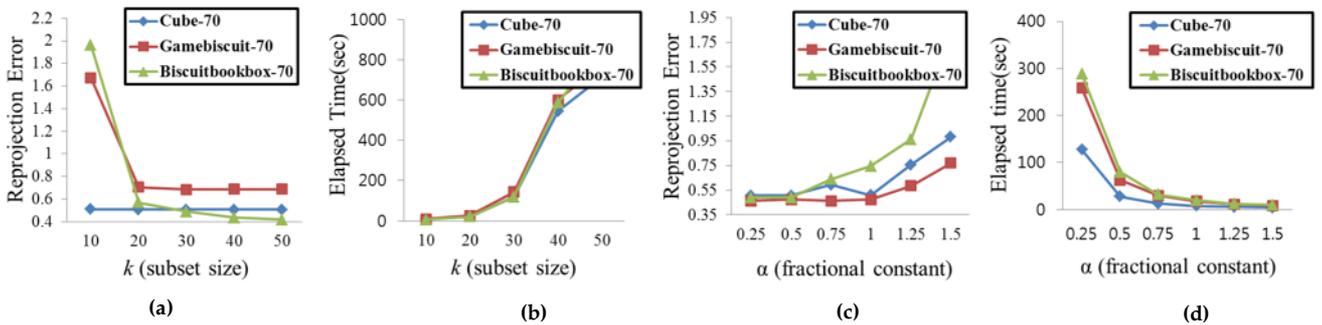

Fig. 8. The effect of parameter $k$ (subset size) and α (fractional constant) on different datasets with 70% of outliers for the affine fundamental matrix estimation with MaxFS-IRL1 method. (a) The re-projection errors obtained from different $k$. (b) The computation time measured from different $k$. (c) The re-projection errors obtained from different α. (d) The computation time measured from different α.

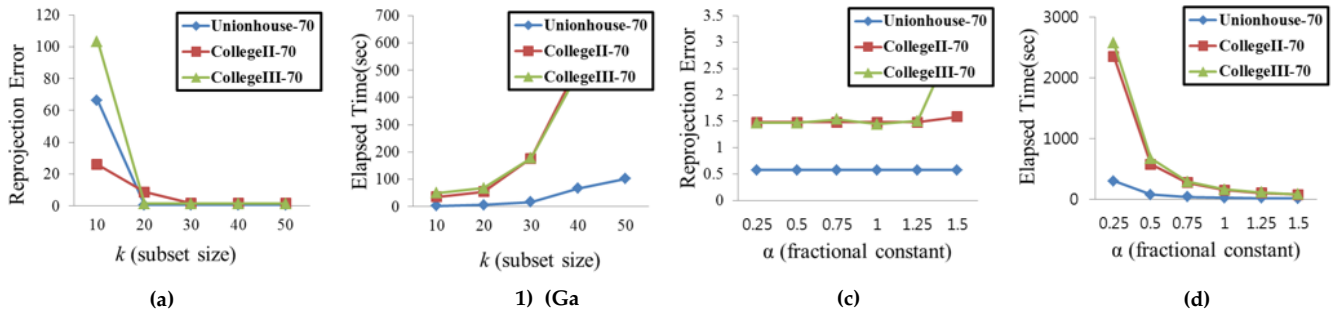

Fig. 7. The effect of parameter $k$ (subset size) and α (fractional constant) on different datasets with 70% of outliers for the homography estimation with MaxFS-IRL1 method. (a) The re-projection errors obtained from different $k$. (b) The computation time measured from different $k$. (c) The re-projection errors obtained from different α. (d) The computation time measured from different α.

Table 2 shows the performance of the algorithms for estimating the affine fundamental matrix for seven datasets (see Fig. 6. (a-g)) with varied outlier ratios. The results show that our algorithm generates high-quality hypotheses with reasonable efficiency for the datasets with high outlier ratios and thus finds all the true structures stably from all of the datasets. In most cases, our method results in the smallest errors on all the test datasets except for the Biscuit book dataset with a relatively low outlier ratio. Since our method performs IRL1 minimization for all of the data in order to generate a hypothesis while other methods generate a hypothesis from minimal subsets, its pling-based approaches as the outlier ratio increases, the maximum errors and standard deviations increase substantially. On the other hand, more robust results were yielded with our MaxFS-IRL1 method since an initial hypothesis generated from the MaxFS was rarely influenced by the outlier ratio.

### 4.2.4 Combination of MaxFS and IRL1

In Figs. 11 and 12, we compared the results using the conventional IRL1 method and using our MaxFS-IRL1 method. We used the CollegeIII dataset. For the conventional IRL1 method, the initial weights were generated from standard $L_1$ minimization from local datasets to compete on par with the MaxFS-IRL1 method. We can see the contribution of each stage from Figs. 11 and 12.

Figures 11(a) and 11(d) show initial fitting results from standard $L_1$ minimization and the MaxFS algorithm for the CollegeIII data with no outliers included. Figures 11(b) and 11(e) show the fitting results after the first reweighted iteration. Figs. 11(c) and 11(f) show the final fitting results from the conventional IRL1 method and the MaxFS-IRL1 method.

Figures 12(a-c) show fitting results from the conventional IRL1 method in each iteration step and Figures 12(d-f) show those from the MaxFS-IRL1 method for the CollegeIII data with 30% of the outliers included.

As appears by these results, when the outlier ratio is low, the standard $L_1$ minimization can provide good results. When there are severe outliers, however, standard $L_1$ minimization frequently fails. To obtain good fitting results using the IRL1 method, it is important to determine good initial weights for the algorithm. Therefore, the MaxFS and the IRL1 algorithms are complementary and we show that their combination is highly effective for the task of robust multiple-structure fitting.

model fitting. For reliable hypothesis generation with reasonable computational efficiency, we employ a MaxFS (maxium feasible subsystem) algorithm, a global optimization technique, only in spatially localized image regions and refine the hypotheses using an IRL1 (re-weighted L1) minimization method. To search out all of the structures thoroughly, the local circular regions spatially overlap with the neighboring ones, and the number of data in each local region is distributed evenly for computational efficiency. The model parameters of major structure are estimated in each local image region, and those of the remainder can be found in one of the neighboring regions. The IRL1 minimization is performed over all the image data to refine initial hypotheses estimated from subsets and to get rid of residual tolerance dependency of the MaxFS algorithm.

Our experiments show that without prior knowledge of the inlier ratio, inlier scale and the number of structures, our method generates consistent hypotheses which are more reliable than the random sampling-based methods as outlier ratios increase.

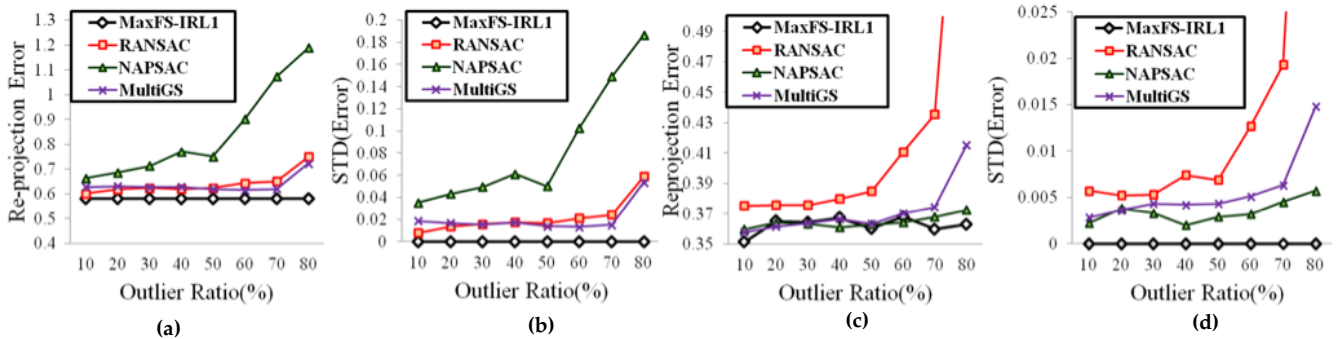

Fig. 10. The performance of four algorithms under different outlier rates. Graphs (a) and (c) show the re-projection errors produced by the four methods on the two test datasts as outlier ratio increases. (b) and (d) show the corresponding standard deviations for the re-projection errors.

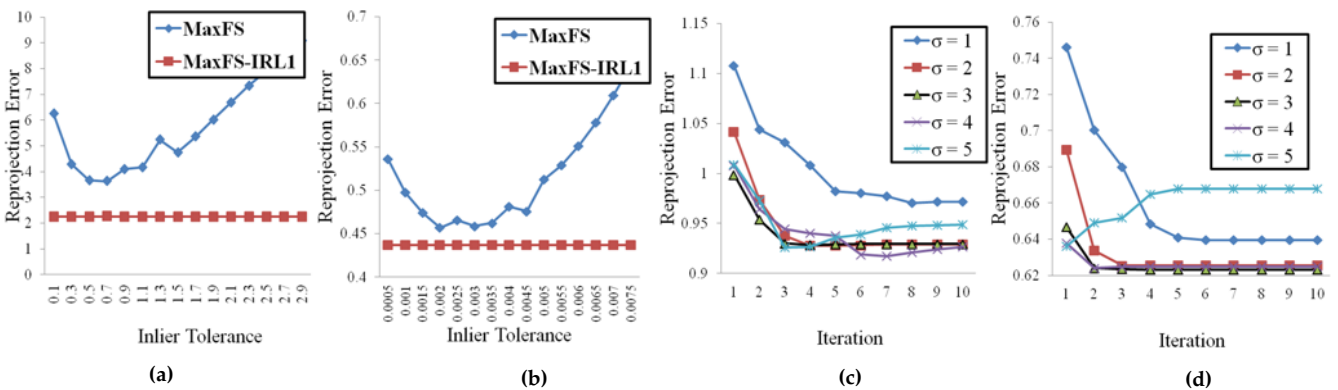

Fig. 9. (a) and (b) show the re-projection errors from the MaxFS algorithm and MaxFS-IRL1 method for one of several structures in the CollegeIII data with 60% of the outliers included and the Carchipscube data with 40% of the outliers included. (c) and (d) show the re-projection errors obtained from different $\sigma$ for each iteration for the Neem data with 50% of the outliers included and the Cubebreadtoychips data with 30% of the outliers included.

## 5 CONCLUSION

We present a new deterministic approach to reliable and consistent hypothesis generation for multiple-structure

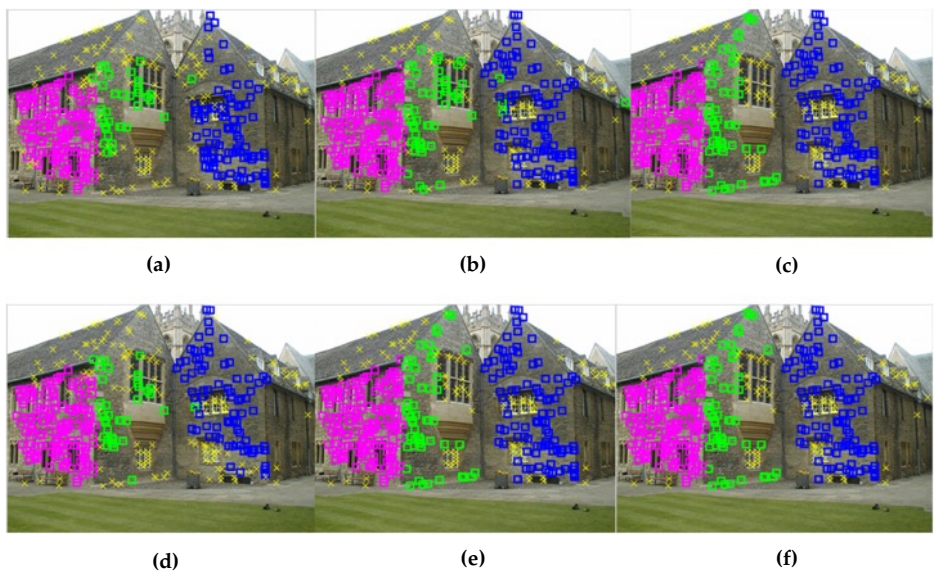

Fig. 11. (a) and (d) show initial fitting results from the standard $L_1$ minimization and the MaxFS algorithm for the CollegeIII data with no outlier included. (b) and (e) show the fitting results after the first reweighted iteration. (c) and (f) show the final fitting results from the conventional IRL1 method and the MaxFS-IRL1 method.

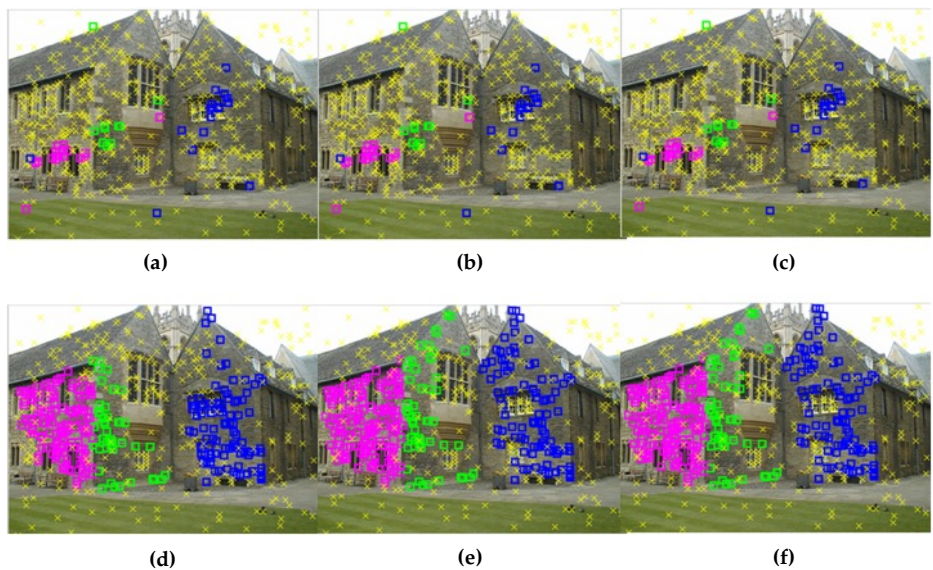

Fig. 12. (a) and (d) show initial fitting results from the standard $L_1$ minimization and the MaxFS algorithm for the CollegeIII data with 30% of the outlier included. (b) and (e) show the fitting results after the first reweighted iteration. (c) and (f) show the final fitting results from the conventional IRL1 method and the MaxFS-IRL1 method.

TABLE 1
PERFORMANCE OF VARIOUS METHODS ON HOMOGRAPHY ESTIMATION FOR THE SEVERAL REAL DATASETS.

| Data | Outlier Ratio[%] | Method | Random | NAPSAC | Multi-GS | MaxFS-IRL1 |
|---|---|---|---|---|---|---|
| **Johnsona** | 10 | Elapsed time [sec] | 15 | 15 | 15 | 13.71 |
| | | Max Error | 1.0871 | 1.5245 | 1.7234 | **0.8512** |
| | | Std | 0.0563 | 0.1351 | 0.1973 | **0** |
| | | L | **12028** | 10926 | 1855 | 24 |
| **Neem** | 20 | Elapsed time [sec] | 10 | 10 | 10 | 9.18 |
| | | Max Error | 1.178 | 1.6387 | 1.0993 | **0.9294** |
| | | Std | 0.0508 | 0.1324 | 0.0352 | **0** |
| | | L | **8386** | 7529 | 1496 | 12 |
| **CollegeIII** | 30 | Elapsed time [sec] | 35 | 35 | 35 | 33.15 |
| | | Max Error | 1.9025 | 2.2848 | 1.9946 | **1.4536** |
| | | Std | 0.0643 | 0.1105 | 0.0983 | **0** |
| | | L | 25043 | 23068 | 2628 | 51 |
| **CollegeII** | 40 | Elapsed time [sec] | 50 | 50 | 50 | 49.19 |
| | | Max Error | 1.5704 | 1.8473 | 1.574 | **1.4851** |
| | | Std | 0.0136 | 0.0668 | 0.0135 | **0** |
| | | L | **35775** | 33819 | 2589 | 35 |
| **Unionhouse** | 50 | Elapsed time [sec] | 10 | 10 | 10 | 7.61 |
| | | Max Error | 0.6507 | 0.8214 | 0.65 | **0.5817** |
| | | Std | 0.0148 | 0.0515 | 0.0165 | **0** |
| | | L | **8559** | 7868 | 2239 | 9 |
| **Elderhalla** | 60 | Elapsed time [sec] | 20 | 20 | 20 | 18.62 |
| | | Max Error | 2.9963 | 2.1821 | 1.9426 | **1.6893** |
| | | Std | 0.2685 | 0.1278 | 0.0711 | **0** |
| | | L | 16922 | 16080 | 2721 | 20 |
| **Sene** | 70 | Elapsed time [sec] | 40 | 40 | 40 | 39.09 |
| | | Max Error | 1.0424 | 0.9275 | 0.6983 | **0.491** |
| | | Std | 0.0953 | 0.1023 | 0.0331 | **0** |
| | | L | **29208** | 28875 | 3061 | 35 |

TABLE 2
PERFORMANCE OF VARIOUS METHODS ON AFFINE FUNDAMENTAL MATRIX ESTIMATION FOR THE SEVERAL REAL DATASETS.

| Data | Outlier Ratio[%] | Method | Random | NAPSAC | Multi-GS | MaxFS-IRL1 |
|---|---|---|---|---|---|---|
| **Biscuitbook** | 20 | Elapsed time [sec] | 20 | 20 | 20 | 21.7 |
| | | Max Error | 1.0147 | **1.0141** | 1.0529 | 1.0591 |
| | | Std | 0.0047 | 0.0056 | 0.0138 | **0** |
| | | L | 20444 | 21041 | 1839 | 108 |
| **Carchipscube** | 30 | Elapsed time [sec] | 20 | 20 | 20 | 17.08 |
| | | Max Error | 0.5542 | 0.5023 | 0.5356 | **0.4927** |
| | | Std | 0.0171 | 0.0036 | 0.0112 | **0** |
| | | L | 29832 | 27005 | 3295 | 74 |
| **Breadcartoychips** | 40 | Elapsed time [sec] | 30 | 30 | 30 | 28.8 |
| | | Max Error | 0.789 | 0.7454 | 0.7344 | **0.7114** |
| | | Std | 0.0205 | 0.0084 | 0.0101 | **0** |
| | | L | 31450 | 28929 | 3023 | 159 |
| **Gamebiscuit** | 50 | Elapsed time [sec] | 35 | 35 | 35 | 35.53 |
| | | Max Error | 0.7196 | 0.6752 | 0.6863 | **0.6744** |
| | | Std | 0.0085 | 0.0038 | 0.0066 | **0** |
| | | L | 31028 | 28927 | 2914 | 178 |
| **Cubetoy** | 50 | Elapsed time [sec] | 30 | 30 | 30 | 28.58 |
| | | Max Error | 0.6469 | 0.6317 | 0.6257 | **0.6076** |
| | | Std | 0.0085 | 0.0058 | 0.0048 | **0** |
| | | L | 24468 | 26301 | 2328 | 194 |
| **Dinobooks** | 60 | Elapsed time [sec] | 40 | 40 | 40 | 40.39 |
| | | Max Error | 2.8726 | 2.3352 | 2.2649 | **2.2078** |
| | | Std | 0.1693 | 0.0567 | 0.04 | **0** |
| | | L | 25307 | 23109 | 2599 | 156 |
| **Cube** | 70 | Elapsed time [sec] | 30 | 30 | 30 | 32.4 |
| | | Max Error | 0.5301 | 0.5486 | 0.5197 | **0.5053** |
| | | Std | 0.0079 | 0.0086 | 0.004 | **0** |
| | | L | 26585 | 27652 | 2558 | 159 |